\documentclass[sn-basic]{sn-jnl}

\usepackage{graphicx}%
\usepackage{multirow}%
\usepackage{amsmath,amssymb,amsfonts}%
\usepackage{amsthm}%
\usepackage{mathrsfs}%
\usepackage[title]{appendix}%
\usepackage{xcolor}%
\usepackage{textcomp}%
\usepackage{manyfoot}%
\usepackage{booktabs}%
\usepackage{algorithm}%
\usepackage{algorithmicx}%
\usepackage{algpseudocode}%
\usepackage{listings}%
\usepackage{tabularx}

\theoremstyle{thmstyleone}%
\theoremstyle{thmstyletwo}%

\theoremstyle{thmstylethree}%

\begin{document}

\title{Motion Forecasting for Autonomous Vehicles: A Survey} 



\author[1]{\sur{Jianxin Shi}}\email{shijx@buaa.edu.cn}
\equalcont{These authors contributed equally to this work.}
\author[1]{\sur{Jinhao Chen}}\email{chenjh24@buaa.edu.cn}
\equalcont{These authors contributed equally to this work.} 
\author*[2]{\sur{Yuandong Wang}}\email{wangyd@cnu.edu.cn}
\author[3]{\sur{Li Sun}}\email{ccesunli@ncepu.edu.cn}
\author[4]{\sur{Chunyang Liu}}\email{liuchunyang@didiglobal.com}
\author[4]{\sur{Wei Xiong}}\email{xiongwei@didiglobal.com}
\author[1]{\sur{Tianyu Wo}}\email{woty@buaa.edu.cn}

\affil[1]{\orgdiv{Beijing Advanced Innovation Center for Big Data and Brain Computing}, \orgname{Beihang University}, \orgaddress{\city{Beijing}, \postcode{100191}, \country{China}}}

\affil*[2]{\orgdiv{Information Engineering College}, \orgname{Capital Normal University}, \orgaddress{\city{Beijing}, \postcode{100048}, \country{China}}}

\affil[3]{\orgdiv{School of Control and Computer Engineering}, \orgname{North China Electric Power University}, \orgaddress{\city{Beijing}, \postcode{102206}, \country{China}}}

\affil[4]{\orgname{Didi Chuxing}, \orgaddress{\city{Beijing}, \postcode{100094}, \country{China}}}






\abstract{In recent years, the field of autonomous driving has attracted increasingly significant public interest. Accurately forecasting the future behavior of various traffic participants is essential for the decision-making of Autonomous Vehicles (AVs). 
In this paper, we focus on both scenario-based and perception-based motion forecasting for AVs.
We propose a formal problem formulation for motion forecasting and summarize the main challenges confronting this area of research. We also detail representative datasets and evaluation metrics pertinent to this field. Furthermore, this study classifies recent research into two main categories: supervised learning and self-supervised learning, reflecting the evolving paradigms in both scenario-based and perception-based motion forecasting. In the context of supervised learning, we thoroughly examine and analyze each key element of the methodology. For self-supervised learning, we summarize commonly adopted techniques. The paper concludes and discusses potential research directions, aiming to propel progress in this vital area of AV technology.}




\maketitle

\section{Introduction}
Motion Forecasting is vital in the functionality of autonomous driving systems. It assists these vehicles in planning their forthcoming actions and mitigates the risk of accidents. 
This survey addresses motion forecasting in autonomous vehicles, focusing on the two main approaches: Scenario-based Motion Forecasting and Perception-based Motion Forecasting.

\textbf{Scenario-based Motion Forecasting} predicts future states of traffic agents (TAs) by analyzing past states and relevant environmental context, such as high-definition maps (HDMaps) and the historical states of surrounding agents (SAs). This approach emphasizes structured, predefined inputs like agents' locations and HDMaps, intentionally excluding raw sensor data like RGB images, LiDAR point clouds, or semantic segmentation maps. By limiting input features to these structured elements, scenario-based forecasting models achieve a focused analysis of the traffic environment and agent interactions.

\textbf{Perception-based Motion Forecasting}, on the other hand, directly utilizes raw perception data, including camera images, LiDAR point clouds, and other sensor outputs, to predict agents' future trajectories. This approach bypasses intermediate feature engineering steps, allowing the model to learn relevant representations directly from raw data. Perception-based methods aim to leverage richer environmental cues, making them suitable for scenarios where comprehensive scene understanding is crucial.

\begin{figure}[H]
    \centering
    \includegraphics[width=\textwidth]{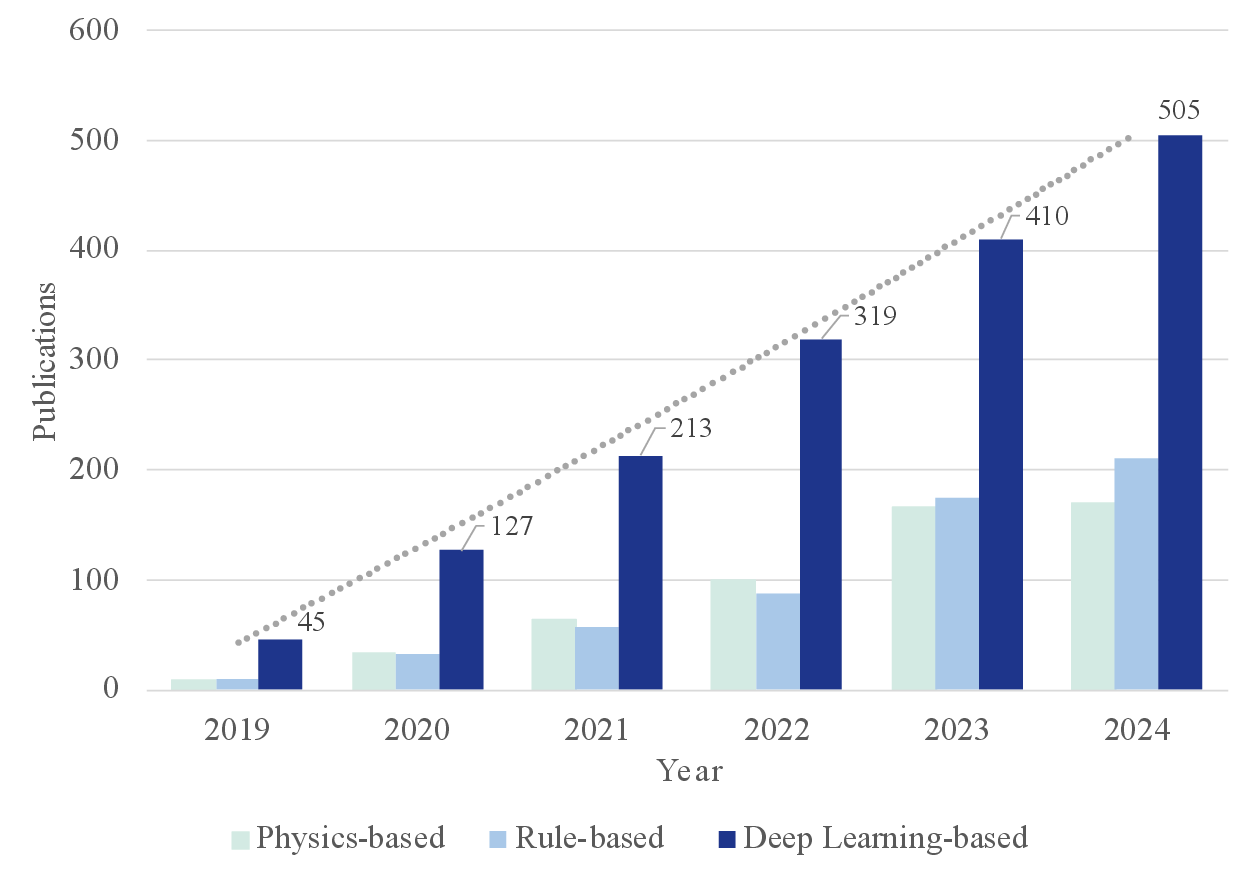}
    \caption{The increasing trend within the research community is evidenced by the growing number of articles on Google Scholar that include the keywords "Autonomous Driving" and "Vehicle Motion Forecasting" published between 2019 and 2024.}
    \label{Figure 1}
\end{figure}

Multiple approaches have been suggested to tackle the prediction problem, including physics-based models, rule-based models, and deep learning-based models. 
Among these, \textbf{physics-based models} \cite{xie2017vehicle} offer a distinct approach by utilizing physical principles to predict vehicle trajectories over the short term. These models incorporate factors such as current position, acceleration, and turn rate to estimate future movements. To elaborate, 
Constant Velocity (CV) models~\cite{scholler2020constant} operate under the assumption that a vehicle maintains its speed in the same direction without acceleration. Similarly, 
Constant Acceleration (CA) models~\cite{polychronopoulos2007sensor} predict movement based on unchanging acceleration in the current direction. 
For scenarios involving both constant speed and constant turn rate, Constant Turn Rate and Velocity (CTRV) models~\cite{lytrivis2008cooperative} are applied. 
Meanwhile, Constant Turn Rate and Acceleration (CTRA) models~\cite{barth2008will} anticipate that a vehicle will maintain both its acceleration and turn rate consistently. Implementing these physics-based approaches is relatively straightforward, requiring minimal computing resources. However, these models have limitations, notably their overlook of environmental factors and interactions with other agents. As a result, while they are efficient for short-term predictions in uncomplicated environments, their applicability is limited by the complexity of the environment and the presence of multiple dynamic agents, underlining the necessity for more sophisticated models in certain contexts. 
\textbf{Rule-based models}~\cite{liu2022probabilistic,qingkai2020lightweight} leverage known traffic rules and human prior knowledge, predicting the future trajectory of vehicles with a structured approach. Initially, models confirm the current state of target agents, a task typically handled by the tracking module that outputs to the prediction module. Following this, the current lane of the vehicle is determined based on its present position and direction. Subsequently, the future lane is predicted according to the vehicle's current state. The final step involves generating the trajectory based on two scenarios: maintaining the current lane or changing lanes. The simplicity and intuitiveness of this method stem from the use of predefined rules, which also contribute to its low computational complexity, ease of adjustment, and reliability. These advantages make rule-based models adapt to various scenarios. However, the inflexibility and limited generalization capabilities of rule-based models pose significant challenges. Since the rules are hard-coded, adapting to new and unseen scenes becomes problematic. Furthermore, capturing and quantifying complex scenes or non-linear interactions is difficult, escalating the complexity of maintaining such a rule-based approach as the number of unforeseen scenarios increases. 
\textbf{Deep learning-based models} have significantly advanced motion forecasting for the long term as Figure~\ref{Figure 1}, leveraging the capabilities of sequential neural networks to extract complex patterns and relationships from extensive datasets. 

These models excel at integrating a wide range of factors, including those related to physics, road conditions, and vehicle interactions, making them exceptionally adaptable to the intricate dynamics of traffic environments. Among the diverse approaches, RNN-based networks \cite{multipath,covernet}, Graph-based networks \cite{vectornet,tnt,densetnt,lanegcn,lanercnn}, and Transformer-based networks \cite{2021scene-transformer,2021AutoBot,2022multi-modal-transformer,2022hivt} stand out, each offering unique strengths in analyzing temporal and spatial data complexities. All these methodologies are based on supervised learning, which has traditionally dominated the field. However, the challenge of acquiring high-precision, labeled trajectory data suitable for autonomous driving prediction has prompted a shift toward innovative solutions. In the last two years, self-supervised learning has emerged as a promising direction in the realm of autonomous vehicles. 
This approach aims to mitigate the scarcity of high-quality data by using strategies such as data augmentation \cite{Pretram,azevedo2022exploiting,li2023pre} to enhance and diversify available datasets. These self-supervised techniques represent a proactive response to the limitations of traditional supervised learning, offering a pathway to refine motion forecasting despite the hurdles posed by data constraints.

\begin{figure}[H]
    \centering
    \includegraphics[width=\textwidth]{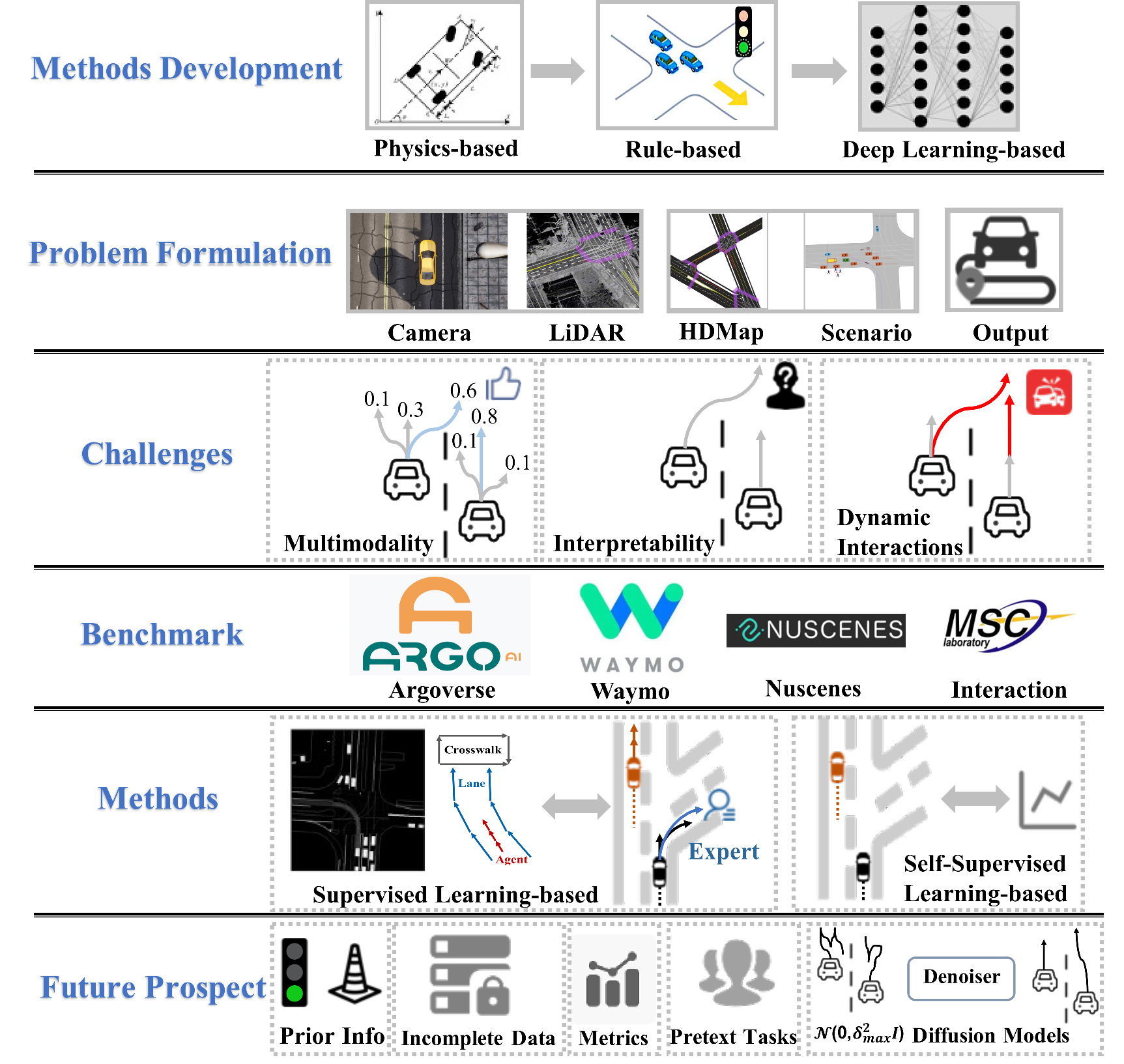}
    \caption{Overview of motion forecasting methodologies, challenges, and future directions for autonomous vehicle trajectory prediction.}
    \label{Figure 2}
\end{figure}

This survey aims to provide a comprehensive review of the latest research in motion forecasting for AVs, covering the general pipelines of both scenario-based and perception-based methods. The structure of the survey is outlined as follows: Section 2 introduces the problem formulation of trajectory prediction, laying out the foundational concepts. Section 3 delves into the main challenges faced in the motion forecasting task, highlighting key areas of difficulty. Section 4 offers a comparison of commonly used motion forecasting datasets and elaborates on the corresponding three levels of evaluation metrics. Sections 5 and 6 are dedicated to discussing the motion forecasting sequence network, with Section 5 focusing on approaches based on supervised learning and Section 6 on those utilizing self-supervised learning. The survey concludes with Section 7, where we present our conclusion and explore potential directions for future research, as Figure~\ref{Figure 2}.


\section{Problem Formulation}
This section introduces some important definitions and formally formulates the problem of motion forecasting. 
\subsection{Definitions}


\subsubsection{Traffic Participants}

\indent\textbf{Target Agents (TAs).} Target Agents represent the objects that are crucial for analysis and prediction in autonomous vehicle systems. Their future behavior or trajectory is of utmost importance for the safe and efficient operation of autonomous vehicles.

\textbf{Ego Agent (EA).} Ego Agent refers to an autonomous vehicle whose behavior is influenced by its surrounding environment. This includes the behavior of TAs, road conditions, and other similar factors.

\begin{figure}[H]
    \centering
    \includegraphics[width=0.5\textwidth]{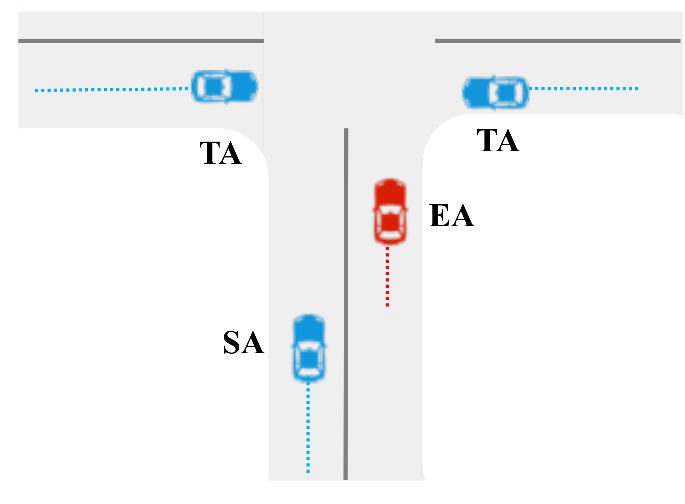}
    \caption{Different types of traffic participants (agents).}
    \label{Figure 3}
\end{figure}

\textbf{Surrounding Agents (SAs).} Surrounding Agents refer to objects such as other vehicles, bicycles, pedestrians, and similar entities that can potentially impact the future behavior of TAs. Different studies use varied criteria for selecting SAs, depending on the specific assumptions underlying their models. 
The visualization of their relationship in the traffic scene is shown in Figure~\ref{Figure 3}. 


\subsubsection{Input Representation}

Motion forecasting can be divided based on different types of input data, each providing unique information that can enhance prediction accuracy. In this section, we review motion forecasting methods categorized by their input types: trajectory and HDMap data, Bird’s Eye View (BEV) representation, and raw perception data. Each category leverages distinct characteristics of the environment and agent dynamics, offering varied advantages in predicting future trajectories.

\textbf{Raw Perception Data.}
Raw perception data serves as an essential component in motion forecasting tasks, providing unprocessed sensor information directly from the environment. In the context of autonomous driving, this data typically comes from sensors like LiDAR, radar, and cameras, offering a richer and more detailed view of the surroundings compared to structured data like trajectories or HD maps. The raw perception data includes point clouds, visual images, and radar reflections, which capture both the static and dynamic aspects of the scene.

A point cloud, for instance, generated by LiDAR sensors, can be formalized as:
\begin{equation}
P = \{(x_{p1}, y_{p1}, z_{p1}, r_{p1}), \dots, (x_{pM}, y_{pM}, z_{pM}, r_{pM})\}
\end{equation} 
where \((x, y, z)\) represents the 3D position of each point in space, and \(r\) is the reflectance value indicating the intensity of the return signal. The number of points, \(M\), may vary based on the environment and the sensor’s resolution.

Similarly, camera images provide 2D pixel information, capturing visual elements such as road signs, obstacles, or pedestrians. This can be represented as a tensor:
\begin{equation}
I = \{(I_{r,g,b})_{w,h}\},
\end{equation}
where \(I_{r,g,b}\) denotes the RGB values for each pixel located at width \(w\) and height \(h\) of the image. Radar data, though less detailed than LiDAR or camera, can detect the velocity of objects with higher accuracy and provides valuable complementary information for forecasting.

Integrating raw perception data into motion forecasting models presents both challenges and opportunities. The high dimensionality and unstructured nature of raw sensor data require sophisticated feature extraction techniques, such as deep learning-based methods, to transform the data into a form suitable for trajectory prediction. Methods like voxelization of point clouds or the extraction of salient features from images are commonly used to handle these raw inputs.

\textbf{Scenario Representation.} For an autonomous vehicle motion prediction problem, the scene representation consists of two parts: High Definition (HD) Map and the states of SAs during past \(T_{obs}\) time.
The mathematical representation of HDMap can be formalized as:
\begin{gather}
    HDMap = \{Lanes=\{(x_{11},y_{11},Attr_1,Lights_1(t)),...,\notag\\
              (x_{l1},y_{l1},Attr_l,Lights_l(t)),...,\notag\\(x_{L1},y_{L1},Attr_L,Lights_L(t))\};\notag\\
              Lights_l(t)=\{state_{t1}(t),...,state_{tT_{obs}}(t)\} \},
\end{gather}
where \((x,y\)) and \(Attr\) represent the position and attribute of the lane segment (such as intersection or not, speed limit or not), respectively. \(L\) is the number of lane segments. \(Lights_l\) represents the change in traffic light status for lane segment $l$, which is a discrete variable containing three elements: "Red", "Yellow", and "Green".

For surrounding agents, their state in the past \(T_{obs}\) time can be formalized as:
\begin{equation}
X_{SAs}=\{X_{1,t-T_{obs}},...,X_{N,t}\},
\end{equation}
where \(X\) represents physical state characteristics of \(N\) surrounding agents in the past \(T_{obs}\) time, such as the position, velocity, heading, etc.

\textbf{BEV Representation.}
Bird’s Eye View (BEV) representations convert raw sensor data, such as LiDAR point clouds or camera images, into a 2D grid format that simplifies the processing and modeling of spatial relationships. This transformation enables motion forecasting models to efficiently capture interactions between agents and their environment, as well as improve the prediction of future trajectories.

The BEV representation can be formalized as:
\begin{equation}
BEV = \{(x_{bev}, y_{bev}, f(x_{bev}, y_{bev}))\},
\end{equation}
where \((x_{bev}, y_{bev})\) are the 2D coordinates in the BEV plane, and \(f(x_{bev}, y_{bev})\) represents various features extracted from the input data, such as occupancy, velocity, or semantic information (e.g., lane markings, drivable areas). The grid size and resolution of the BEV representation determine the trade-off between computational efficiency and prediction accuracy.

BEV representations have gained popularity due to their ability to encode complex spatial information and facilitate the modeling of multi-agent interactions. Many state-of-the-art motion forecasting models leverage BEV to predict the future paths of both the ego vehicle and surrounding agents in a unified manner.

\subsubsection{Prediction Types}


\textbf{Marginal Trajectory Prediction.} In the marginal prediction case, the trajectories of TAs do not affect each other, the prediction goal can be formalized as:
\begin{equation}
Output\_Marginal_{T_{pred}}=p(s_1)p(s_2)...p(s_n),
\end{equation}
where \(p(s_i)\) represents the predicted trajectory distribution of agent \(i\) in the future, which is related to the historical states of agent \(i\), the historical states of agent \(j\), and the HDMap.

\begin{figure}[H]
    \centering
    \includegraphics[width=\textwidth]{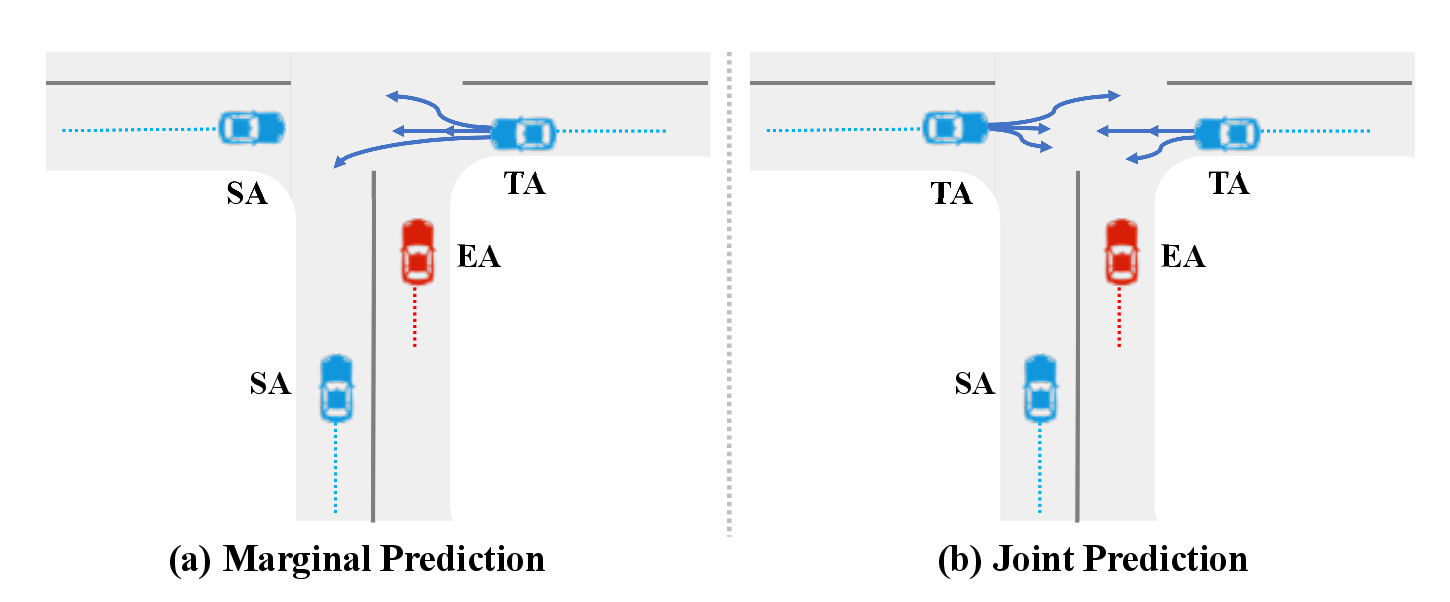}
    \caption{The relationship between marginal prediction and joint multi-agent prediction. The main difference between them is that the former has a single $TA$; The latter has multiple $TAs$.}
    \label{Figure 4}
\end{figure}

\textbf{Joint Multi-Agent Prediction.}Unlike marginal trajectory prediction, joint multi-agent prediction is more complex and essential as it requires predicting the future trajectories of multiple vehicles while considering their mutual interactions and influences, as shown in Figure~\ref{Figure 4}. This prediction method integrates the states and intentions of all vehicles, better reflecting the dynamic changes in real traffic environments. 

In the joint prediction case, compared with marginal prediction, the interdependencies between TAs' future behavior can be captured, and the prediction goal is:
\begin{equation}
Output\_Joint_{T_{pred}}=p(s_1,s_2,...,s_n),
\end{equation}
Compared to the marginal prediction, \(p(s_i)\) is also related to the future prediction trajectory of agent \(j\). 

\subsection{Problem Formulation}

\begin{figure}[H]
    \centering
    \includegraphics[width=\textwidth]{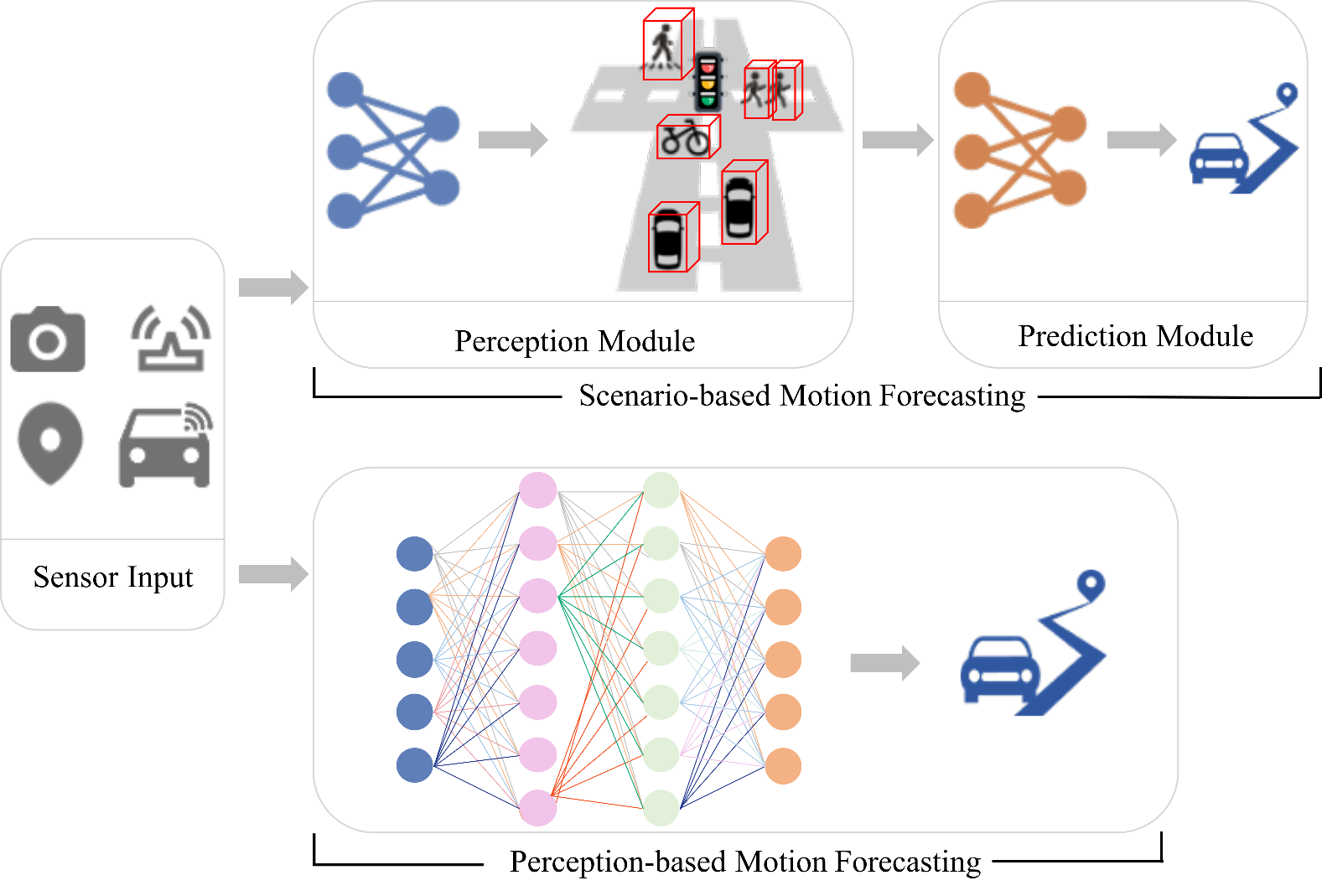}
    \caption{Comparison of scenario-based and perception-based motion forecasting pipelines in autonomous driving.}
    \label{Figure 5}
\end{figure}

Motion forecasting pipelines can generally be categorized into two main approaches based on the data flow and processing stages: \textbf{Scenario-based Motion Forecasting} and \textbf{Perception-based Motion Forecasting}. 
The pipelines for these two approaches are illustrated in Figure \ref{Figure 5}.

\subsubsection{Scenario-based Motion Forecasting}
The primary objective of motion forecasting is to predict the future states of TAs, considering their past states as well as the surrounding traffic scenes. These traffic scenes encompass various elements, including HDMaps, past states of SAs, and other relevant factors. In contrast to survey \cite{teeti2022vision-based-survey} focusing on vision-based prediction models in AVs, the scenario-based motion forecasting model restricts its input features to agents' location information and HDMaps. These models specifically exclude other types of data such as RGB images, Lidar point clouds, semantic segmentation maps, and any other additional data. Considering these specific inclusions and exclusions, the overall input for these models can be formalized as:
\begin{equation}
Input_{T_{obs}}=\{X_{TAs},HDMap,X_{SAs},Others\},
\end{equation}
where \(Others\) represents the other information such as traffic congestion of the current lane, lighting status of SAs, and the planned trajectory of EA, etc.

The forecasting process then predicts the future trajectories of the target agents based on this structured input. The predicted trajectory for a target agent can be formalized as:
\begin{equation}
\hat{X}_{TA} = \{(x_{ta,t+1}, y_{ta,t+1}), (x_{ta,t+2}, y_{ta,t+2}), ..., (x_{ta,t+T}, y_{ta,t+T})\},
\end{equation}
where \((x_{ta,t+i}, y_{ta,t+i})\) represents the position of the target agent at each future time step \(t+i\), and \(T\) is the prediction horizon.


By focusing on structured inputs such as HD maps and past agent states, scenario-based motion forecasting aims to leverage the explicit representations of the environment and agent dynamics to achieve accurate trajectory predictions.

\subsubsection{Perception-based Motion Forecasting} 
Perception-based motion forecasting refers to the approach of directly predicting the future trajectories of agents from raw perception data, such as camera images, LiDAR point clouds, and other raw sensor data. without the need for artificially designed intermediate features or steps. 

Mathematically, the end-to-end motion forecasting process can be described as learning a function \(f\), which maps the raw perception data \(Z_{t}\) (e.g., LiDAR point clouds or camera images) to the future trajectories \(X_{t+T}\) of the agents:
\begin{equation}
f: Z_{t} \to X_{t+T},
\end{equation}
where \(Z_{t} = \{Z_{LiDAR}, Z_{camera}, Z_{radar}, ...\}\) represents the raw sensor data at time \(t\), and \(X_{t+T}\) is the predicted trajectory of the target agent(s) over the future time horizon \(T\).

The predicted trajectory \(X_{t+T}\) can be further formalized as a sequence of future states over \(T\) time steps:
\begin{equation}
X_{t+T} = \{X_{t+1}, X_{t+2}, ..., X_{t+T}\},
\end{equation}
where each future state \(X_{t+i}\) typically includes information such as position, velocity, and heading of the agent at time \(t+i\). The learning objective for the model is to minimize the prediction error between the ground truth future trajectory \(X^{*}_{t+T}\) and the predicted trajectory \(X_{t+T}\), using a loss function such as mean squared error (MSE):
\begin{equation}
\mathcal{L} = \sum_{i=1}^{T} || X_{t+i} - X^{*}_{t+i} ||^2.
\end{equation}

By training the model end-to-end with this objective, it learns to capture the intricate dynamics of the environment and agent interactions, directly from the raw perception data, enabling accurate and robust motion forecasting.

By bypassing intermediate steps like handcrafted feature extraction or external map information, these models can potentially avoid the error propagation often seen in staged pipelines. Furthermore, end-to-end approaches allow the model to learn implicit representations of the environment, agents, and their interactions from raw data, potentially leading to richer and more nuanced predictions. 

\section{Challenges}
In the domain of autonomous driving, motion forecasting for TAs is very supportive of the EA's next move. However, accurate motion forecasting for TAs remains a challenging task due to the complexity and flexible traffic environment.

\textbf{Fusion of road information.} 
To advance the autonomous driving field, researchers utilize more detailed features and achieve centimeter-level accuracy for vehicle behavior prediction by constructing HDMaps. 
These HDMaps provide rich contextual information, such as lane boundaries, traffic signs, and road geometry, which are crucial for making precise and reliable predictions. However, the absence of unified standards in HDMaps' data formats and content poses significant challenges. How to establish data alignment and association between HDMaps and agents' trajectories, and effectively integrate this information in vehicle behavior prediction is a huge challenge.

\textbf{Dynamic interactions between different vehicles.} 
The influence of road environments on vehicle behavior is static, whereas the interaction between SAs and TAs is dynamic and uncertain, posing significant challenges in capturing this complex interplay. For instance, a vehicle's decision to turn right at an intersection involves interacting with the static environment of the right-turn lane. However, the dynamic and variable nature of interactions between SAs and TAs adds complexity, making the analysis and interpretation of these interaction patterns significantly more challenging.

\textbf{Multimodality of vehicle behavior.} 
In autonomous driving, understanding the behaviors of TAs and SAs is critical due to their inherent multimodality, meaning a single historical trajectory can lead to multiple potential future trajectories. 
The combination of an agent's trajectory with road information provides valuable insights into the driver's style, especially their familiarity with specific road sections. A deeper analysis of the agent's historical movement patterns enables the identification of various possible future behaviors. Consequently, an effective motion forecasting module in an autonomous driving system should be able to recognize all these potential future behaviors. This capability is essential to ensure the system operates reliably and safely.

\textbf{Lack of interpretability.}
Many existing motion forecasting models adopt data-driven methodologies to learn trajectory distributions. While these approaches can achieve high levels of accuracy by leveraging large datasets, they often result in a lack of interpretability in the decision-making processes of traffic participants. This black-box nature makes it challenging to understand or explain why a model predicts certain behaviors, which is crucial for both safety and trust in autonomous systems. Moreover, the reliance on purely data-driven techniques can lead to overfitting specific scenarios or biases present in the data, potentially reducing the generalizability and robustness of the models in diverse and dynamic real-world environments. As a result, there is an increasing need for approaches that not only predict motion accurately but also offer clear insights into the underlying factors influencing these predictions. 


\section{Training and Evaluation}
In this section, we introduce several open datasets utilized in motion forecasting for autonomous vehicles, as well as commonly used metrics.
\subsection{Datasets}
A summary of public motion forecasting datasets for AVs is shown in Table~\ref{tab:plain}. Details are described below.
 
Argoverse \cite{2019argoverse1}, developed by Argo AI,  offers a comprehensive collection of urban driving scenarios, extensively annotated for research in 3D tracking and forecasting. 
Building on the foundation of its predecessor, Argoverse 2 motion forecasting dataset \cite{2023argoverse2} expands upon the original dataset with a larger volume of data, including 250,000 scenarios, each scenario provides a local vector map and 11 seconds of trajectory data (recorded at 10 Hz). The first 5 seconds of data represent the observation window, while the following 6 seconds correspond to the forecast horizon. 
Waymo Open Motion Dataset \cite{2021waymo}, created by Waymo, encompasses a wide array of driving conditions and scenarios, making it one of the most comprehensive resources available. It comprises over 100,000 scenes, each 20 seconds long and sampled at 10 Hz. This amounts to more than 570 hours of unique data, covering over 1750 km of roadways.
The Interaction dataset \cite{zhan2019interaction} is distinct in its focus on interactive driving scenarios, particularly those involving complex urban intersections and roundabouts. The Interaction dataset includes a rich collection of real-world driving scenarios with a significant number of vehicle trajectories, totaling over 41,000 across all categories and covering more than 990 minutes of driving behavior. 
The nuScenes dataset \cite{2020nuscenes} is notable for its extensive coverage of varied driving environments, encompassing 1,000 scenes that span across various weather conditions and times of the day collected in Boston and Singapore.
This rich collection of sensor data not only enhances the realism of the driving scenarios but also provides invaluable insights for the development and refinement of motion forecasting models, making nuScenes a pivotal tool for researchers in the field.

\begin{table*}
    \centering
    \caption{Widely Used Datasets for AVs in motion forecasting.}
    \label{tab:plain}
    \begin{tabularx}{\textwidth}{l|lllX}
        \toprule
        Dataset   & Year  & Length & Scenes & Typical Models \\
        \midrule
        Interaction     &2019  & -     & -
        &IDE-Net \cite{IDE-Net}, ADAPT \cite{adapt}, Traj-MAE \cite{Traj-MAE}, T4P~\cite{park2024t4p}
        \\
        Waymo     & 2021     & 9s        & 104k    & Scene-transformer~\cite{2021scene-transformer}, M2I~\cite{sun2022m2i}, MTR~\cite{MTR}, Wayformer~\cite{nayakanti2023wayformer}, HDGT~\cite{jia2023hdgt}, JFP~\cite{luo2023jfp}, GameFormer~\cite{huang2023gameformer}, MotionLM~\cite{seff2023motionlm}, MotionDiffuser~\cite{jiang2023motiondiffuser}, MTR\texttt{++}~\cite{shi2024mtr++}, T4P~\cite{park2024t4p}
        \\
        nuScenes  & 2020     & 8s        & 41k    &  Trajectron\texttt{++}~\cite{salzmann2020trajectron++}, MHA-JAM~\cite{messaoud2021trajectory}, AgentFormer \cite{2021agentformer}, AutoBot~\cite{2021AutoBot}, PreTraM~\cite{Pretram}, Forecast-PEFT~\cite{wang2024forecast}, T4P~\cite{park2024t4p}
        \\
        Argoverse  & 2019     & 5s        & 324k        & Vectornet \cite{vectornet}, LaneGCN \cite{lanegcn}, Scene transformer \cite{2021scene-transformer}, TNT \cite{tnt}, DenseTNT \cite{densetnt}, LaneRCNN \cite{lanercnn}, \cite{2021mm-transformer}, HOME \cite{home}, GOHOME \cite{gohome}, Hivt \cite{2022hivt}, SSL-Lanes \cite{ssl-lanes}, Traj-MAE \cite{Traj-MAE}, ADAPT~\cite{adapt}, Forecast-PEFT~\cite{wang2024forecast}, SmartRefine~\cite{zhou2024smartrefine} \\
        Argoverse 2 & 2023     & 11s        & 250k       & Forecast-MAE \cite{Forecast-mae}, POP \cite{POP}, SEPT~\cite{SEPT}, QCNet~\cite{qcnet}, QCNext~\cite{qcnext}, SmartRefine~\cite{zhou2024smartrefine}, Forecast-PEFT~\cite{wang2024forecast}
        \\
        \bottomrule
    \end{tabularx}%
    
\end{table*}
\subsection{Evaluation Metrics}
Standardized evaluation settings and commonly used metrics are essential for a data-driven approach to obtain quantitative results. 
The quantitative results allow different models to compare with each other from diverse perspectives. The metrics used frequently in motion forecasting can be summarized from the following three levels.

\textbf{Geometry-level Metric.}
Geometric measurement \cite{2019argoverse1} serves as a crucial index for assessing the similarity between predicted and actual trajectories, effectively representing accuracy. 

\textit{1) The minimum Average Displacement Error (minADE)}: ADE measures the averaged L2 distance of all future timesteps. MinADE is used to evaluate multimodal trajectory prediction, which measures the L2 distance in meters between the best-predicted trajectory and the ground truth trajectory averaged over all future timesteps. The best-predicted trajectory is defined as the one that has the minimum endpoint error. 
For K trajectory samples of each agent, the formula of the \(minADE_{K}\) is:
\begin{equation}
minADE_{K}=\frac{min_{k=1}^{K}\sum_{t=T_{obs}}^{T_{pred-1}}\left \| \hat{y}_i^{t,(k)} -{y}_i^{t} \right \|  }{T_{pred}-T_{obs}},
\end{equation}
where \(\hat{y}_i^{t,(k)}\) represents the predicted states of target agent \(i\) at time \(t\) in the \(k\)-th sample, and \({y}_i^{t}\) is the corresponding ground truth.

\textit{2) The minimum Final Displacement Error (minFDE):} FDE measures the endpoint L2 distance of all future timesteps. MinFDE is used to evaluate multimodal trajectory prediction, which measures the error between the best-predicted trajectory and the ground truth trajectory at the final future time step. 
For K trajectory samples of each agent, the formula of the \(minFDE_{K}\) is:
\begin{equation}
minFDE_{K}=min_{k=1}^{K}\left \| \hat{y}_i^{T_{pred-1},(k)} -{y}_i^{T_{pred-1}} \right \|,
\end{equation}
where \(\hat{y}_i^{T_{pred-1},(k)}\) denotes the predicted states of target agent \(i\) at final future time \(T_{pred-1}\) in the \(k\)-th sample, and \({y}_i^{T_{pred-1}}\) is the corresponding ground truth.

\textit{3) Miss Rate (MR):} The number of scenarios in which all predicted trajectories deviate by more than 2.0 meters from the ground truth, as measured by endpoint error.

\textbf{Probabilistic-level Metric.}
The different versions of Negative Log Likelihood (NLL) \cite{salzmann2020trajectron++} can be used as a probability measurement, comparing the distribution of the generated trajectories against the ground truth to evaluate uncertainty, especially for multimodal output distributions. As an example, consider the NLL for Laplace distribution \cite{2022hivt,qcnet}: 
\begin{equation}
NLL_{Laplace}={\textstyle \sum_{t=T_{obs}}^{T_{pred-1}}} \log{P(y_{i}^{t}|\hat{y}_{i}^{t},\hat{b}_{i}^{t})} ,
\end{equation} 
where $P(\cdot|\cdot)$ represents the probability density function of Laplace distribution, $\hat{b}_{i}^{t}$ is the uncertainties of the best-predicted trajectory for agent $i$.

\textbf{Task-level Metric.}
Task measurement \cite{task-metric} is used to evaluate the impact of trajectory prediction on the downstream planning module. The planning-informed (PI)
versions of accuracy-based metrics (e.g., piADE and piFDE) can be instantiated using:
\begin{equation}
PI-Metric =\frac{1}{|\mathcal{A}|} \sum_{a \in \mathcal{A}} f\left(a,\left|\nabla_{\hat{\mathbf{s}}^{(t: T)}} c\right|\right) \cdot \operatorname{Metric}\left(\hat{\mathbf{s}}_a^{(t: T)}, \mathbf{s}_a^{(t: T)}\right),
\end{equation}
where \(\operatorname{Metric}\) represents the accuracy-based metrics (e.g., ADE and FDE) and the function \(f\) enables the application of various schemes for assigning weights. For each agent \(a\in\mathcal{A}\), \(\mathbf{s}_a^{(t: T)}\) and \(\hat{\mathbf{s}}_a^{(t: T)}\) are the predicted positions in the next \(T\) timestamps and the ground truth positions, respectively.

\section{Model Architecture} 
\subsection{Supervised Learning-based Architecture}
\begin{figure}[H]
    \centering
    \includegraphics[width=\textwidth]{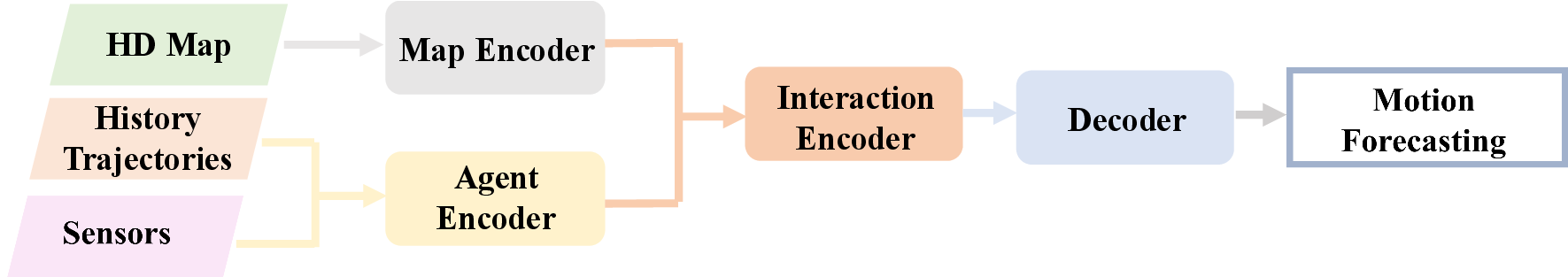}
    \caption{SL-based architecture for motion forecasting.}
    \label{Figure 6}
\end{figure}

Figure~\ref{Figure 6} shows the general pipeline of Supervised Learning-based (\textbf{SL-based}) architecture employed in motion forecasting tasks, where the Encoder and Decoder can be any practicable network components, such as attention mechanism, GNNs, and transformers. 
As vehicle motions or trajectories are typically temporal-spatial data, we introduce SL-based architecture from temporal-spatial encoding and decoding, respectively. 

\subsubsection{Temporal-Spatial Encoding}
During the encoding, both temporal and spatial features need to be extracted. Spatially, the agent-to-lane and agent-to-agent interactions could be modeled. Temporally, the dynamics of these interactions along with time are captured through designed components.

\textbf{Rasterized-based Encoder.} Rasterized-based approaches rasterize the map information and agent states of each timestep into an image. Then, a scenario can be modeled into a time series of images. Existing methods employ CNNs to learn effective representations from spatial and temporal perspectives \cite{intentnet,multipath,covernet}. For instance, 

One of the early typical explorations in this area involved the FAF model~\cite{luo2018fast}, which converts LiDAR point cloud data into a top-down bird’s-eye view. This data is then processed by a target detection network equipped with a CNN-based multi-frame information fusion module to extract spatio-temporal correlations from the perception sequence. To enhance the model's performance, a trajectory regression loss function was incorporated into the detection head, allowing for the end-to-end optimization of object localization and future trajectory prediction. FAF not only enables joint 3D object detection and trajectory prediction from LiDAR point cloud data, but also demonstrates that incorporating the prediction task improves the accuracy of object detection. 

Based on FaF~\cite{luo2018fast}, IntentNet~\cite{intentnet} enhances vehicle intent detection by generating additional outputs within a unified detection and prediction framework, where intent is defined as short-term motion states and lane-related actions. 

Similarly, Multipath \cite{multipath} employs CNNs to extract features from raster images, followed by predicting probabilities across K-predefined anchor trajectories and calculating regressed offsets from the anchor states. 
CoverNet \cite{covernet} also utilizes CNNs for feature extraction from raster images. 
This approach offers a comprehensive pixel-level visual context, enabling models to discern and assimilate diverse visual patterns directly. However, it is noteworthy that converting traffic scenes into pixel format might result in losing some intrinsic topological information. Consequently, recent research has predominantly utilized graph-based and attention-based methods, as discussed below. 

\textbf{Graph-based Encoder.}
Graph-based encoding approaches generally initiate with vectorized representations of HDMaps and agents, integrating additional attribute data into these vectors. 
These approaches ensure the comprehensive retention of information about HDMaps and agents within the graph data structure. 

Graph-based neural networks are utilized for feature extraction following the vectorized representation of traffic scene components. These include Graph Attention Networks (GAT), which focus on interactive feature extraction, and Graph Convolutional Networks (GCN), aimed at extracting road topological features. Vectornet \cite{vectornet} introduces an innovative hierarchical graph network structure. In its first level, this network aggregates spatial information based on polylines, while the second level is dedicated to modeling complex interrelationships between polylines. Simultaneously, additional tasks are proposed to augment the second-level graph's capability in capturing interactions between various agents and other elements represented in the HDMap. 
Following this, LaneGCN \cite{lanegcn} utilizes an attention mechanism to capture complex interactions. In contrast, LaneRCNN \cite{lanercnn} employs a global graph-based module specifically designed to learn the dynamic relationships between different agents in the system. 
When it comes to end-to-end motion forecasting frameworks, SpAGNN~\cite{casas2020spagnn} introduces a graph neural network into the joint detection and prediction framework, leveraging graph models from trajectory data research to account for the interactions between multiple agents, thereby generating more accurate predicted trajectories.


\textbf{Attention-based Encoder.}
The advent of transformers has marked a significant breakthrough in multi-modal prediction within recent years. Its unique attention-based module can fully explore the interaction between vehicles in highly dynamic scenarios and effectively model the multi-modal distribution of trajectories.
\cite{messaoud2021trajectory} introduced an innovative approach that utilizes multi-head attention to create a joint representation of static scenes and surrounding agents. Each attention head in this model is designed to represent a potential interaction pattern between the target agent and the combined context features.
AgentFormer \cite{2021agentformer} designed a spatio-temporal transformer along with better multi-modal properties. It incorporates all observable agent sequences in a scene to perform complex cross-sequence input processing, facilitated by end-to-end training.
\cite{2021mm-transformer} features three separate stacked Transformer models that aggregate historical tracks, road information, and interaction details.
Scene transformer \cite{2021scene-transformer} unifies motion prediction and goal-conditioned motion prediction. It employs diverse agent/time masking strategies and alternating types of attention between different modalities to capture interactions effectively.
\cite{2022multi-modal-transformer} proposes a multi-modal attention transformer encoder. This model adapts the multi-head attention mechanism to accommodate multi-modal attention, where each predicted trajectory is influenced by a unique attention mode.
AutoBot \cite{2021AutoBot} proposes a transformer-based network utilizing an axial attention mechanism to learn the temporal and spatial correlations among agents and road topology.
Furthermore, HiVT \cite{2022hivt}, QCNet \cite{qcnet}, and QCNext \cite{qcnext} focus on modeling both local and global contexts in a translation and rotation invariant transformer network. 
GameFormer \cite{huang2023gameformer} leverages hierarchical game theory and transformer architectures to tackle the challenge of interaction prediction for autonomous vehicles. Through a structured learning process and iterative refinement of predictions, the model achieves superior accuracy and performance, significantly advancing the state-of-the-art in the field.

\subsubsection{Visual Encoding}

Some models leveraging a single neural network to jointly address both detection and prediction tasks, known as "perception-based motion forecasting," have been proposed to enhance computational efficiency and accuracy~\cite{luo2018fast}. These approaches typically begin by quantizing the 3D world into a voxel grid and assigning a binary indicator to each voxel to denote whether it is occupied. Convolutional operations are then performed on this grid to extract feature information and predict future behavior. 
While these methods incorporate the characteristics of sensor detection and prediction, they fail to utilize the rich temporal information associated with traffic agents. PnPNet~\cite{liang2020pnpnet} overcomes this limitation by encoding long-term historical data through a combination of online tracking and the extraction of trajectory-level agent representations, resulting in enhanced performance across all tasks. 
AffiniPred~\cite{weng2022affinipred}
and ~\cite{zhang2022trajectoryforecastingdetectionuncertaintyaware} perform implicit data association by
using detections and their affinity matrices as inputs instead of working on past trajectories. These studies have a special focus on the tracking error while tackling a subset of imperfections with an adversarial scene or object generation. Differently, ~\cite{xu2024motionforecastingrealworldperception} focus on understanding the impacts of real-world inputs from various state-of-the-art perception
methods on the different motion forecasting paradigms. 

While many approaches integrate sensor detection and prediction, they often overlook the temporal richness of traffic agents. ViP3D~\cite{gu2023vip3dendtoendvisualtrajectory} addresses this by encoding agent dynamics with 3D queries, capturing both visual and motion features from multi-view images through cross-attention. The queries are stored in a memory bank to track agents over time, and interact with HD maps for final trajectory prediction, ensuring end-to-end differentiability.
UniAD~\cite{hu2023planningorientedautonomousdriving} further builds on these ideas by offering a unified approach to autonomous driving, integrating perception, prediction, and planning into a single framework. It leverages task-specific queries and transformer modules to synchronize key tasks like tracking, motion forecasting, and planning, ensuring consistent outputs across them.


\subsubsection{Trajectory Decoding}
With the representations containing both spatial-temporal features and interaction features between traffic agents, a decoder needs to be devised to generate multi-modal future trajectories. There are two decoding ways, anchor-conditioned and anchor-free. 

\textbf{Anchor-Conditioned Decoding.} The anchor-conditioned decoding approach typically incorporates prior knowledge from the dataset as an input component of the network, facilitating the generation of multi-modal trajectories, essentially conditional probabilities. Based on these various prior anchors, the final output trajectory can be constrained within a set. However, the effectiveness of this method largely depends on the quality and relevance of these predefined anchors. Depending on the type of anchor, this type of decoding approach further includes: goal-based decoder, heatmap-based decoder, and intention-based decoder. 

Goal-based Decoder. 
Recently, goal-based multi-trajectory prediction methods have proven to be effective. These methods operate on the principle that the endpoint carries most of the uncertainty of the trajectory, so they first predict the agent's target and then further complete the corresponding full trajectory for each target. The final target position is obtained by classifying and regression the predefined sparse anchor points. 
TNT \cite{tnt} defines an anchor point as a location sampled on the centerline of a lane segment. The offsets on the x and y axes are predicted based on the candidate anchor point, and the expected end point is obtained by combining the offset with the anchor. The trajectory is then finalized based on this endpoint.
LaneRCNN \cite{lanercnn} uses lane segments as anchors and predicts a goal for each lane segment. 
\cite{lane-graph-traversals} devises a sampling strategy aimed at predicting the potential future paths of vehicles. This method involves generating various paths through different sampling techniques. Each path is then associated with a specific motion mode, characterized by an introduced latent variable typically modeled as a normal distribution.
DenseTNT \cite{densetnt} directly outputs a set of trajectories from dense goal candidates. However, the method's reliance on an online optimization strategy for these dense goal candidates is highly computationally demanding. 
Building on the concepts from TNT and DenseTNT, ADAPT \cite{adapt} proposes an innovative method for the concurrent prediction of trajectories for all agents within a scene. 
This is achieved through dynamic weight learning. The process begins with the prediction of a potential set of endpoints. Each endpoint is then refined by predicting an offset, and finally, the complete trajectories are determined based on these endpoints.
However, such anchor-conditioned decoders rely heavily on the density of the goal. In scenarios where the density of goals is high, the required computational power can exceed practical limits.

Heatmap-based Decoder. 
HOME \cite{home} introduces an approach that utilizes probabilistic heatmaps as the output format for trajectory prediction. This methodology employs a full convolutional model, but it is constrained by the limitations of a fixed image size.
Building upon the foundations laid by HOME, GOHOME \cite{gohome} advances this concept by proposing a motion prediction framework that is predicated entirely on graph manipulation optimization. 
THOMAS \cite{thomas} adopts the same graph encoder, sampling algorithm, and full trajectory generation mechanism as GOHOME. However, it incorporates an efficient layered heatmap process, which is scalable for making concurrent predictions of multiple bodies. In addition, a scenario consistency module has been added to recombine the marginal prediction output into a federated prediction.

Intention-based Decoder. 
IntentNet \cite{intentnet} manually defined several intentions for autonomous vehicles, such as turning left and changing lanes, and learned a separate motion predictor for each intention. 
Multipath \cite{multipath} considers the uncertainty of trajectory comes from two parts: the uncertainty of intention and the uncertainty of control. The uncertainty of intention is managed through a fixed set of future state-sequence anchors that represent various modes of trajectory distribution. Once an intention is identified, the control aspect also introduces uncertainty about future predictions. This uncertainty is modeled as a normal distribution at each time step and is represented by a parameterized offset relative to the anchor trajectory. Thus, a single forward inference can yield the future distribution of multimodal trajectories.
CoverNet \cite{covernet} approaches trajectory prediction as a classification challenge within a discrete set of trajectory clusters. These clusters are designed to encompass the possible state space while adhering to the constraints of vehicle dynamics. Additionally, the trajectory clusters are tailored to align with the current kinematic state of the vehicle, such as not turning around at high speed, making small turns, etc.

\textbf{Anchor-Free Decoding. }
Compared with the anchor-conditioned decoding approach, the ordinary non-anchor approach does not set the anchor prior based on data in advance and takes the anchor prior as input, but directly outputs the forecast trajectory from the decoder. However, this approach lacks spatial prior information from the data, the prediction results tend to learn the modes with the highest frequency, while the mode with the lowest frequency is not sufficiently learned, and its accuracy will decline in long-term prediction tasks.
Therefore, a novel paradigm of learnable anchor decoding has been proposed, which takes into account the advantages of both anchor-based and non-anchor solutions. 
MTR~\cite{MTR} designed a motion query pair that combines global intention localization and local movement refinement, implementing a "global first, local second" approach, achieving better prediction results. 
Multipath\texttt{++} \cite{multipath++} learn anchor embeddings as an integral part of the overall model training, rather than being pre-set data elements. This method establishes a direct correlation between the potential space of anchor embeddings and the multimodal output of the mixed Gaussian distribution.
QCNet \cite{qcnet} devises a two-stage prediction strategy. The first stage encompasses a coarse prediction phase, where a recurrent and anchor-free proposal module is employed to generate adaptive trajectory anchors. Following this, the second stage involves an anchor-based module that refines these proposed trajectory anchors, enhancing the accuracy and specificity of the predictions.
Building on the foundations of QCNet, QCNext \cite{qcnext} recognizes that QCNet is primarily tailored for marginal trajectory prediction. To address this limitation, QCNext proposes a new DETR-like decoder that can capture future social interactions for multi-agent joint prediction tasks.


\subsection{Self-Supervised Learning-based Architecture}
\begin{figure}[b]
    \centering
    \includegraphics[width=\textwidth]{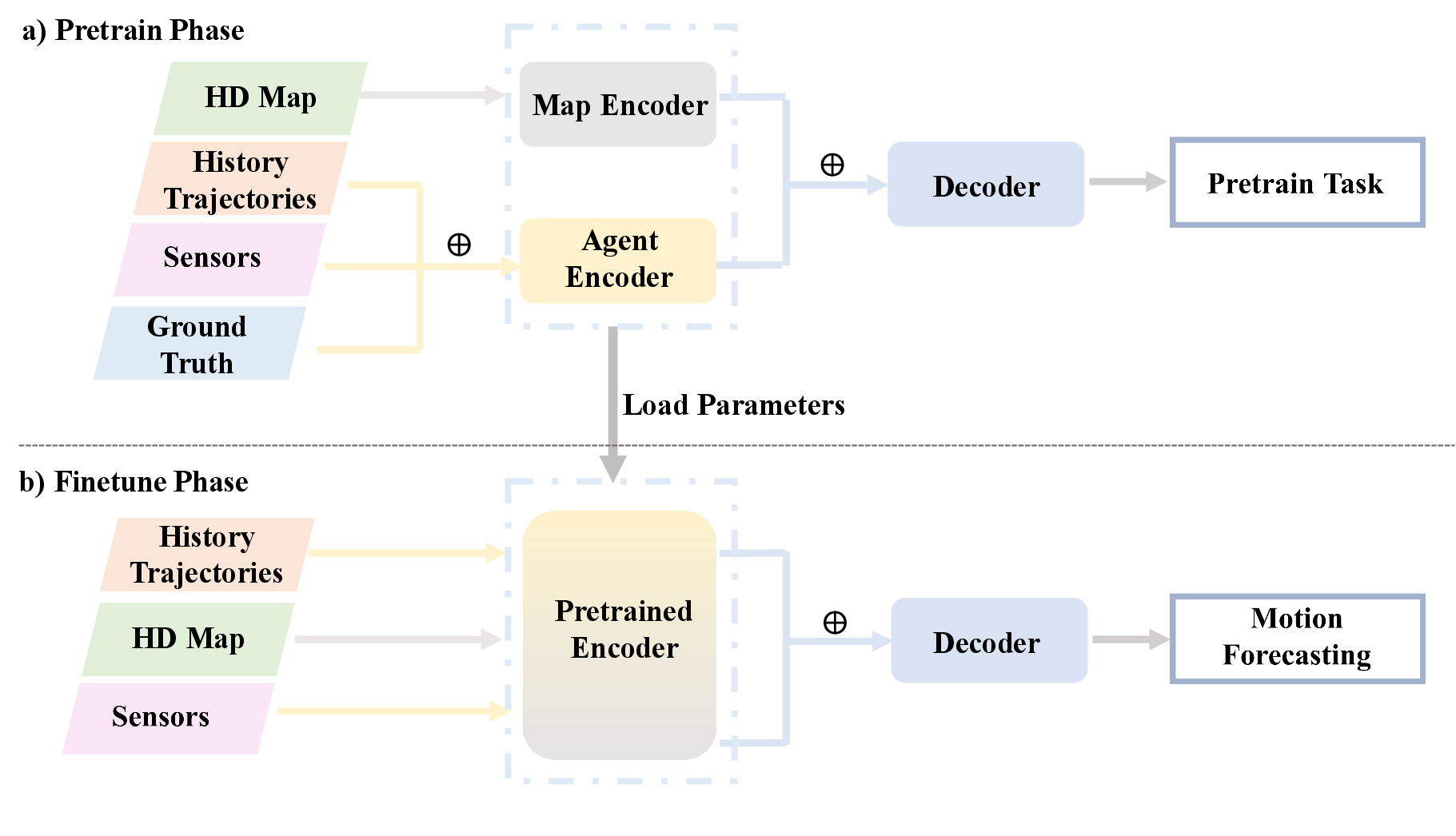}
    \caption{SSL-based architecture for motion forecasting.}
    \label{Figure 7}
\end{figure}

Self-Supervised Learning (\textbf{SSL}) is widely applied in natural language processing and computer vision, benefiting from the availability of large-scale unlabeled data. 
There are already studies proving that SSL is effective in helping models learn a more comprehensive representation for downstream tasks.
Hence, SSL has begun to be explored in motion forecasting, aiming for more transferable and robust representation learning.
Figure~\ref{Figure 7} illustrates the general pipeline of SSL-based Architecture in motion forecasting.

\subsubsection{Preliminary Exploration}
Vectornet's \cite{vectornet} introduction of a graph-based completion auxiliary task represents a pioneering exploration in motion forecasting for autonomous driving, utilizing a self-supervised learning approach. This innovation serves as a foundational step in applying self-supervision techniques in this field, potentially paving the way for future advancements. 
PreTraM \cite{Pretram} considers that the limited availability of trajectory data restricts SSL's applicability in motion forecasting. To mitigate this, it introduces a method for generating supplementary rasterized map patches, derived from localized areas of comprehensive HDMaps, for the training of a robust map encoder through contrastive learning. Furthermore, PreTraM innovatively employs a pre-training strategy for both map and trajectory encoders, which involves the pairing of batches of training instances to effectively enhance the encoders' ability to learn the complex relationship between maps and trajectories. 
SSL-Lanes \cite{ssl-lanes} proposes a comprehensive set of four pretraining tasks focusing on one specific input modality, including lane masking. These tasks encompass intersection distance calculation, maneuver classification, and success/failure classification. SSL-Lanes illustrates that the strategic design of pretext tasks can significantly enhance performance. This enhancement is primarily achieved through the extraction of more complex features from existing data, eliminating the need for additional datasets.

\subsubsection{Data Augmentation} 
\cite{azevedo2022exploiting} develops an innovative method to augment and diversify the limited motion data available for trajectory prediction. This approach involves an auxiliary task utilizing HDMaps to generate potential trajectories for traffic agents. These trajectories are constructed using synthetic speeds and by leveraging the interconnectedness of graph nodes. The overarching objective is to achieve precise motion prediction, a goal that remains constant from the pre-training stage through to the fine-tuning phase. Nevertheless, the implementation of this approach requires intricate modeling of agent positions and the generation of synthetic velocities, a process that becomes particularly challenging when dealing with non-annotated data.
\cite{li2023pre} considers that the methodology developed by \cite{azevedo2022exploiting}, which entails generating pseudo trajectories that rigorously adhere to lane structures for pre-training purposes, introduces an innovative data generation strategy. This strategy is designed to generate realistic synthetic trajectories, bridging the domain gap between synthetic and real-world data.
\cite{Road-Barlow} employs a dual-phase training approach for several transformer models aimed at motion prediction. The pre-training phase involves the use of basic map data, to align with the Barlow Twins paradigm by learning embeddings for augmented yet similar views of this map data. The fine-tuning phase is distinct, utilizing samples annotated with historical data of traffic agents to specifically enhance the model's accuracy in motion forecasting tasks.
\subsubsection{MAE-based Approach} 
Following the significant progress in image-based self-supervised representation learning, Masked AutoEncoder (MAE) \cite{he2022mask-mae} has attracted considerable interest across various fields. The core mechanism of this approach is to mask part of the input data, followed by the application of an autoencoder structure aimed at reconstructing the masked tokens, thereby enhancing learning efficiency. 

Traj-MAE \cite{Traj-MAE} first presents a novel and efficient masked trajectory autoencoder specifically for self-supervised trajectory prediction, designs two independent mask-reconstruction tasks on trajectories and road map input to train its trajectory and map encoder separately. The research further explores various masking strategies, including both social and temporal aspects, to facilitate the trajectory encoder in capturing latent semantic information from multiple perspectives. However, the methodology exhibits a significant limitation: the spatial relationship between agents and roads is insufficiently emphasized during the pretraining phase. Forecast-MAE \cite{Forecast-mae,wang2024forecast} devises a scene reconstruction task employing a novel masking strategy during the pre-training phase. This involves masking the historical trajectory of some agents, the future trajectory of others, and applying a random masking method for lanes. This approach enables the model to effectively capture agents' behavior patterns, road environment features, and their interactions.

\subsubsection{Auxiliary task Design}
POP \cite{POP} identifies the critical challenge of performance degradation when utilizing prediction algorithms in the case of insufficient observations. To mitigate this, it develops a reconstruction branch, which focuses on reconstructing the obscured historical elements of partially observed datasets. This reconstruction is facilitated through an advanced masking procedure coupled with a reconstruction head. 
SEPT \cite{SEPT} integrates three self-supervised masking-reconstruction tasks focusing on scene inputs, which encompass both agents’ trajectories and the road network. 
These tasks are designed to pre-train the scene encoder, enabling it to effectively capture the kinematics inherent within a trajectory, comprehend the spatial structure of the road network, and recognize interactions between roads and agents. 

\subsubsection{Language Modeling Methods}

Inspired by the success of large language models in addressing complex tasks through model scaling, several approaches~\cite{sun2024largetrajectorymodelsscalable}~\cite{jia2024amp}~\cite{seff2023motionlm} discretizes continuous trajectories into sequences of motion tokens, representing agent actions as selections from a finite vocabulary. This design allows reformulating trajectory generation problems as a unified sequence modeling task, aligning them with other sequence modeling problems such as language modeling.


STR~\cite{sun2024largetrajectorymodelsscalable} arranges all components of motion prediction and planning into a single sequence, including map information, past trajectories of other road users, future states, etc. The core of STR is a causal transformer backbone, specifically using the GPT-2 model. This choice enables easy scaling of the model size and incorporation of recent developments in language modeling.
AMP~\cite{jia2024amp} employs tailored position encodings to handle complex spatial-temporal relations, including relative spatial position encoding, temporal position encoding, and rotary position encoding (RoPE). This design allows AMP to unify input and output representations in an ego-centric coordinate system and perform autoregressive prediction in a GPT-style manner, while addressing the unique challenges of the autonomous driving domain.
MotionLM~\cite{seff2023motionlm} consists of: (a) A scene encoder that processes multimodal inputs including roadgraph elements, traffic light states, and features describing road agents and their recent histories. (b) A trajectory decoder that autoregressively generates sequences of discrete motion tokens for multiple agents. It is trained using a maximum likelihood objective over multi-agent action sequences. At inference time, the model can sample diverse trajectory rollouts, which are then aggregated to produce a set of representative joint modes for evaluation. This architecture allows MotionLM to capture complex interactions between agents while maintaining temporal causality in its predictions.

\section{Conclusion and Prospect}
In this paper, we present a comprehensive review of the recent advancements in motion forecasting for autonomous vehicles. We begin by introducing the formulation of motion forecasting and then move on to an overview of diverse, widely-utilized datasets. This is followed by a detailed explanation of evaluation metrics specifically designed for motion forecasting. State-of-the-art prediction models have made significant strides, employing advanced techniques such as attention mechanisms, GNNs, transformers, and self-supervised architectures. Despite these technological advances, the field still faces substantial challenges. Understanding motion forecasting is pivotal for autonomous driving, as it greatly enhances the interpretation of road scenarios, thereby playing a crucial role in improving the safety standards of future autonomous driving technologies.

\textbf{Fusion of more prior information.}
Recent research has integrated HDMaps into motion forecasting models. This integration specifically involves incorporating lane information to ensure predicted trajectories are aligned with the road topology. In real-world scenarios, other factors also play a crucial role. These include traffic light status, various traffic signs, and additional elements that influence the movement and interactions of traffic participants. However, many current methodologies tend to overlook these aspects. This oversight results in limitations in the mechanistic understanding of motion forecasting models. As a consequence, the impact of traffic indication information on the behavior of traffic participants remains a relatively underexplored area.

\textbf{Model robustness under incomplete scenario.}
The majority of motion forecasting models operate under the presumption that all observational data is fully accessible. 
However, this assumption rarely holds true in real-world traffic scenarios, where various factors can limit the availability and quality of observational data. Sensor constraints, such as limited range or resolution, and environmental factors, such as weather conditions or low lighting, can impair the sensors' ability to capture all relevant information. Additionally, object occlusion, where vehicles, pedestrians, or other objects block the sensors' line of sight, can result in significant portions of the scene going unobserved. 
In these cases, the EA may only partially observe TAs and SAs, leading to gaps in the data that are critical for accurate motion forecasting. This incomplete observation can degrade the performance of conventional models, which often rely on the assumption of comprehensive data to make accurate predictions. Therefore, there is a pressing need to develop more advanced motion forecasting models that can function robustly even when faced with incomplete observation data. 

\textbf{Alignment of evaluation metrics.}
Despite significant advancements by joint perception-to-forecasting models such as ViP3D~\cite{gu2023vip3dendtoendvisualtrajectory} and UniAD~\cite{hu2023planningorientedautonomousdriving}, these models have not been directly compared to established pure motion forecasting models. This lack of direct comparison is primarily due to inherent differences in their methodological approaches and evaluation criteria. To bridge this gap, it is crucial to develop an adapted evaluation protocol that considers the cascading impact of upstream errors in the perception-to-forecasting pipeline. Implementing such a protocol would enable a more balanced and informative comparison.

\textbf{Exploration of novel pretext tasks.}
The robust representational learning and transferability exhibited by pre-train and fine-tune paradigms in the fields of computer vision and natural language processing have inspired recent works to incorporate these approaches into motion forecasting for autonomous vehicles. These works typically involve designing various reconstruction tasks as pretext tasks and subsequently fine-tuning the model for downstream motion forecasting. Thus, the exploration of novel pretext tasks within the self-supervised learning domain presents promising avenues for further advancements in motion forecasting. 

\textbf{Diffusion Models for Controllable Multi-Agent Trajectory Prediction.}
Recent research~\cite{jiang2023motiondiffuser} has introduced diffusion models to the realm of multi-agent trajectory prediction, emphasizing controllability and realistic inter-agent interactions. This framework enables modeling a joint distribution of agent trajectories, allowing the generation of collision-free, contextually adaptive predictions in complex environments. Despite its promise, current implementations are limited by simplified assumptions and lack comprehensive contextual factors, such as diverse traffic scenarios or dynamic road elements. Exploring ways to integrate richer contextual data and agent-specific behavior could significantly enhance the versatility and accuracy of these models. This emerging direction highlights diffusion models as a powerful yet underexplored tool in achieving robust, adaptable multi-agent motion forecasting.

\newpage

\bibliography{a_final_submit}

\begin{thebibliography}{80}
\providecommand{\natexlab}[1]{#1}
\providecommand{\url}[1]{{#1}}
\providecommand{\urlprefix}{URL }
\providecommand{\doi}[1]{\url{https://doi.org/#1}}
\providecommand{\eprint}[2][]{\url{#2}}
 \bibcommenthead

\bibitem[{Aydemir et~al(2023)Aydemir, Akan, and G{\"u}ney}]{adapt}
Aydemir G, Akan AK, G{\"u}ney F (2023) Adapt: Efficient multi-agent trajectory prediction with adaptation. In: Proceedings of the IEEE/CVF International Conference on Computer Vision, pp 8295--8305

\bibitem[{Azevedo et~al(2022)Azevedo, Gilles, Sabatini, and Tsishkou}]{azevedo2022exploiting}
Azevedo C, Gilles T, Sabatini S, et~al (2022) Exploiting map information for self-supervised learning in motion forecasting. arXiv preprint arXiv:221004672

\bibitem[{Barth and Franke(2008)}]{barth2008will}
Barth A, Franke U (2008) Where will the oncoming vehicle be the next second? In: 2008 IEEE Intelligent Vehicles Symposium, IEEE, pp 1068--1073

\bibitem[{Bhattacharyya et~al(2023)Bhattacharyya, Huang, and Czarnecki}]{ssl-lanes}
Bhattacharyya P, Huang C, Czarnecki K (2023) Ssl-lanes: Self-supervised learning for motion forecasting in autonomous driving. In: Conference on Robot Learning, PMLR, pp 1793--1805

\bibitem[{Caesar et~al(2020)Caesar, Bankiti, Lang, Vora et~al}]{2020nuscenes}
Caesar H, Bankiti V, Lang AH, et~al (2020) nuscenes: A multimodal dataset for autonomous driving. In: Proceedings of the IEEE/CVF conference on computer vision and pattern recognition, pp 11621--11631

\bibitem[{Carion et~al(2020)Carion, Massa, Synnaeve, Usunier, Kirillov, and Zagoruyko}]{detr2020end}
Carion N, Massa F, Synnaeve G, et~al (2020) End-to-end object detection with transformers. In: European conference on computer vision, Springer, pp 213--229

\bibitem[{Casas et~al(2018)Casas, Luo, and Urtasun}]{intentnet}
Casas S, Luo W, Urtasun R (2018) Intentnet: Learning to predict intention from raw sensor data. In: Conference on Robot Learning, PMLR, pp 947--956

\bibitem[{Casas et~al(2020)Casas, Gulino, Liao, and Urtasun}]{casas2020spagnn}
Casas S, Gulino C, Liao R, et~al (2020) Spagnn: Spatially-aware graph neural networks for relational behavior forecasting from sensor data. In: 2020 IEEE International Conference on Robotics and Automation (ICRA), IEEE, pp 9491--9497

\bibitem[{Chai et~al(2019)Chai, Sapp, Bansal, and Anguelov}]{multipath}
Chai Y, Sapp B, Bansal M, et~al (2019) Multipath: Multiple probabilistic anchor trajectory hypotheses for behavior prediction. arXiv preprint arXiv:191005449

\bibitem[{Chang et~al(2019)Chang, Lambert, Sangkloy, Singh, Bak et~al}]{2019argoverse1}
Chang MF, Lambert J, Sangkloy P, et~al (2019) Argoverse: 3d tracking and forecasting with rich maps. In: Proceedings of the IEEE/CVF conference on computer vision and pattern recognition, pp 8748--8757

\bibitem[{Chen et~al(2023)Chen, Wang, Shao, Liu, Hao et~al}]{Traj-MAE}
Chen H, Wang J, Shao K, et~al (2023) Traj-mae: Masked autoencoders for trajectory prediction. arXiv preprint arXiv:230306697

\bibitem[{Cheng et~al(2023)Cheng, Mei, and Liu}]{Forecast-mae}
Cheng J, Mei X, Liu M (2023) Forecast-mae: Self-supervised pre-training for motion forecasting with masked autoencoders. In: Proceedings of the IEEE/CVF International Conference on Computer Vision, pp 8679--8689

\bibitem[{Coifman and Li(2017)}]{2017NGSIM}
Coifman B, Li L (2017) A critical evaluation of the next generation simulation (ngsim) vehicle trajectory dataset. Transportation Research Part B: Methodological 105:362--377

\bibitem[{Cui et~al(2019)Cui, Radosavljevic, Chou, Lin, Nguyen, Huang, Schneider, and Djuric}]{cui2019multimodaltrajectorypredictionsautonomous}
Cui H, Radosavljevic V, Chou FC, et~al (2019) Multimodal trajectory predictions for autonomous driving using deep convolutional networks. \urlprefix\url{https://arxiv.org/abs/1809.10732}, \eprint{1809.10732}

\bibitem[{Deo et~al(2022)Deo, Wolff, and Beijbom}]{lane-graph-traversals}
Deo N, Wolff E, Beijbom O (2022) Multimodal trajectory prediction conditioned on lane-graph traversals. In: Conference on Robot Learning, PMLR, pp 203--212

\bibitem[{Ettinger et~al(2021)Ettinger, Cheng, Caine, Liu, Zhao, Pradhan et~al}]{2021waymo}
Ettinger S, Cheng S, Caine B, et~al (2021) Large scale interactive motion forecasting for autonomous driving: The waymo open motion dataset. In: Proceedings of the IEEE/CVF International Conference on Computer Vision, pp 9710--9719

\bibitem[{Gao et~al(2020)Gao, Sun, Zhao, Shen, Anguelov, Li, and Schmid}]{vectornet}
Gao J, Sun C, Zhao H, et~al (2020) Vectornet: Encoding hd maps and agent dynamics from vectorized representation. In: Proceedings of the IEEE/CVF Conference on Computer Vision and Pattern Recognition, pp 11525--11533

\bibitem[{Geiger et~al(2013)Geiger, Lenz, Stiller, and Urtasun}]{2013kitti}
Geiger A, Lenz P, Stiller C, et~al (2013) Vision meets robotics: The kitti dataset. The International Journal of Robotics Research 32(11):1231--1237

\bibitem[{Gilles et~al(2021{\natexlab{a}})Gilles, Sabatini, Tsishkou, Stanciulescu, and Moutarde}]{thomas}
Gilles T, Sabatini S, Tsishkou D, et~al (2021{\natexlab{a}}) Thomas: Trajectory heatmap output with learned multi-agent sampling. arXiv preprint arXiv:211006607

\bibitem[{Gilles et~al(2021{\natexlab{b}})Gilles, Sabatini, Tsishkou et~al}]{home}
Gilles T, Sabatini S, Tsishkou D, et~al (2021{\natexlab{b}}) Home: Heatmap output for future motion estimation. In: 2021 IEEE International Intelligent Transportation Systems Conference (ITSC), IEEE, pp 500--507

\bibitem[{Gilles et~al(2022)Gilles, Sabatini, Tsishkou, Stanciulescu, and Moutarde}]{gohome}
Gilles T, Sabatini S, Tsishkou D, et~al (2022) Gohome: Graph-oriented heatmap output for future motion estimation. In: 2022 international conference on robotics and automation (ICRA), IEEE, pp 9107--9114

\bibitem[{Girgis et~al(2021)Girgis, Golemo, Codevilla, Weiss, D'Souza et~al}]{2021AutoBot}
Girgis R, Golemo F, Codevilla F, et~al (2021) Latent variable sequential set transformers for joint multi-agent motion prediction. arXiv preprint arXiv:210400563

\bibitem[{Gu et~al(2021)Gu, Sun, and Zhao}]{densetnt}
Gu J, Sun C, Zhao H (2021) Densetnt: End-to-end trajectory prediction from dense goal sets. In: Proceedings of the IEEE/CVF International Conference on Computer Vision, pp 15303--15312

\bibitem[{Gu et~al(2023)Gu, Hu, Zhang, Chen, Wang, Wang, and Zhao}]{gu2023vip3dendtoendvisualtrajectory}
Gu J, Hu C, Zhang T, et~al (2023) Vip3d: End-to-end visual trajectory prediction via 3d agent queries. \urlprefix\url{https://arxiv.org/abs/2208.01582}, \eprint{2208.01582}

\bibitem[{He et~al(2022)He, Chen, Xie, Li, Doll{\'a}r, and Girshick}]{he2022mask-mae}
He K, Chen X, Xie S, et~al (2022) Masked autoencoders are scalable vision learners. In: Proceedings of the IEEE/CVF conference on computer vision and pattern recognition, pp 16000--16009

\bibitem[{Houston et~al(2021)Houston, Zuidhof, Bergamini, Ye, Chen, Jain, Omari, Iglovikov, and Ondruska}]{2021lyft}
Houston J, Zuidhof G, Bergamini L, et~al (2021) One thousand and one hours: Self-driving motion prediction dataset. In: Conference on Robot Learning, PMLR, pp 409--418

\bibitem[{Hu et~al(2023)Hu, Yang, Chen, Li, Sima, Zhu, Chai, Du, Lin, Wang, Lu, Jia, Liu, Dai, Qiao, and Li}]{hu2023planningorientedautonomousdriving}
Hu Y, Yang J, Chen L, et~al (2023) Planning-oriented autonomous driving. \urlprefix\url{https://arxiv.org/abs/2212.10156}, \eprint{2212.10156}

\bibitem[{Huang et~al(2022)Huang, Mo, and Lv}]{2022multi-modal-transformer}
Huang Z, Mo X, Lv C (2022) Multi-modal motion prediction with transformer-based neural network for autonomous driving. In: 2022 International Conference on Robotics and Automation (ICRA), IEEE, pp 2605--2611

\bibitem[{Huang et~al(2023)Huang, Liu, and Lv}]{huang2023gameformer}
Huang Z, Liu H, Lv C (2023) Gameformer: Game-theoretic modeling and learning of transformer-based interactive prediction and planning for autonomous driving. In: Proceedings of the IEEE/CVF International Conference on Computer Vision, pp 3903--3913

\bibitem[{Ivanovic and Pavone(2021)}]{task-metric}
Ivanovic B, Pavone M (2021) Rethinking trajectory forecasting evaluation. arXiv preprint arXiv:210710297

\bibitem[{Jia et~al(2021)Jia, Sun, Tomizuka, and Zhan}]{IDE-Net}
Jia X, Sun L, Tomizuka M, et~al (2021) Ide-net: Interactive driving event and pattern extraction from human data. IEEE Robotics and Automation Letters 6(2):3065--3072

\bibitem[{Jia et~al(2022)Jia, Sun, Zhao, Tomizuka, and Zhan}]{jia2022multi}
Jia X, Sun L, Zhao H, et~al (2022) Multi-agent trajectory prediction by combining egocentric and allocentric views. In: Conference on Robot Learning, PMLR, pp 1434--1443

\bibitem[{Jia et~al(2023)Jia, Wu, Chen, Liu, Li, and Yan}]{jia2023hdgt}
Jia X, Wu P, Chen L, et~al (2023) Hdgt: Heterogeneous driving graph transformer for multi-agent trajectory prediction via scene encoding. IEEE transactions on pattern analysis and machine intelligence

\bibitem[{Jia et~al(2024)Jia, Shi, Chen, Jiang, Liao, He, and Yan}]{jia2024amp}
Jia X, Shi S, Chen Z, et~al (2024) Amp: Autoregressive motion prediction revisited with next token prediction for autonomous driving. arXiv preprint arXiv:240313331

\bibitem[{Jiang et~al(2023)Jiang, Cornman, Park, Sapp, Zhou, Anguelov et~al}]{jiang2023motiondiffuser}
Jiang C, Cornman A, Park C, et~al (2023) Motiondiffuser: Controllable multi-agent motion prediction using diffusion. In: Proceedings of the IEEE/CVF Conference on Computer Vision and Pattern Recognition, pp 9644--9653

\bibitem[{Krajewski et~al(2018)Krajewski, Bock, Kloeker, and Eckstein}]{2018highd}
Krajewski R, Bock J, Kloeker L, et~al (2018) The highd dataset: A drone dataset of naturalistic vehicle trajectories on german highways for validation of highly automated driving systems. In: 2018 21st international conference on intelligent transportation systems (ITSC), IEEE, pp 2118--2125

\bibitem[{Lan et~al(2023)Lan, Jiang, Mu, Chen, Li, Zhao, and Li}]{SEPT}
Lan Z, Jiang Y, Mu Y, et~al (2023) Sept: Towards efficient scene representation learning for motion prediction. arXiv preprint arXiv:230915289

\bibitem[{Li et~al(2023)Li, Zhao, Xu, Tang, Li, Ding, Tomizuka, and Zhan}]{li2023pre}
Li Y, Zhao SZ, Xu C, et~al (2023) Pre-training on synthetic driving data for trajectory prediction. arXiv preprint arXiv:230910121

\bibitem[{Liang et~al(2020{\natexlab{a}})Liang, Yang, Hu, Chen, Liao et~al}]{lanegcn}
Liang M, Yang B, Hu R, et~al (2020{\natexlab{a}}) Learning lane graph representations for motion forecasting. In: Computer Vision--ECCV 2020: 16th European Conference, Glasgow, UK, August 23--28, 2020, Proceedings, Part II 16, Springer, pp 541--556

\bibitem[{Liang et~al(2020{\natexlab{b}})Liang, Yang, Zeng, Chen, Hu, Casas, and Urtasun}]{liang2020pnpnet}
Liang M, Yang B, Zeng W, et~al (2020{\natexlab{b}}) Pnpnet: End-to-end perception and prediction with tracking in the loop. In: Proceedings of the IEEE/CVF Conference on Computer Vision and Pattern Recognition, pp 11553--11562

\bibitem[{Liu et~al(2022)Liu, Luo, Zhong, Li, Huang, and Xiong}]{liu2022probabilistic}
Liu J, Luo Y, Zhong Z, et~al (2022) A probabilistic architecture of long-term vehicle trajectory prediction for autonomous driving. Engineering 19:228--239

\bibitem[{Liu et~al(2021)Liu, Zhang, Fang, Jiang, and Zhou}]{2021mm-transformer}
Liu Y, Zhang J, Fang L, et~al (2021) Multimodal motion prediction with stacked transformers. In: Proceedings of the IEEE/CVF Conference on Computer Vision and Pattern Recognition, pp 7577--7586

\bibitem[{Luo et~al(2018)Luo, Yang, and Urtasun}]{luo2018fast}
Luo W, Yang B, Urtasun R (2018) Fast and furious: Real time end-to-end 3d detection, tracking and motion forecasting with a single convolutional net. In: Proceedings of the IEEE conference on Computer Vision and Pattern Recognition, pp 3569--3577

\bibitem[{Luo et~al(2023)Luo, Park, Cornman, Sapp, and Anguelov}]{luo2023jfp}
Luo W, Park C, Cornman A, et~al (2023) Jfp: Joint future prediction with interactive multi-agent modeling for autonomous driving. In: Conference on Robot Learning, PMLR, pp 1457--1467

\bibitem[{Lytrivis et~al(2008)Lytrivis, Thomaidis, and Amditis}]{lytrivis2008cooperative}
Lytrivis P, Thomaidis G, Amditis A (2008) Cooperative path prediction in vehicular environments. In: 2008 11th International IEEE Conference on Intelligent Transportation Systems, IEEE, pp 803--808

\bibitem[{Messaoud et~al(2021)Messaoud, Deo, Trivedi, and Nashashibi}]{messaoud2021trajectory}
Messaoud K, Deo N, Trivedi MM, et~al (2021) Trajectory prediction for autonomous driving based on multi-head attention with joint agent-map representation. In: 2021 IEEE Intelligent Vehicles Symposium (IV), IEEE, pp 165--170

\bibitem[{Nayakanti et~al(2023)Nayakanti, Al-Rfou, Zhou, Goel, Refaat, and Sapp}]{nayakanti2023wayformer}
Nayakanti N, Al-Rfou R, Zhou A, et~al (2023) Wayformer: Motion forecasting via simple \& efficient attention networks. In: 2023 IEEE International Conference on Robotics and Automation (ICRA), IEEE, pp 2980--2987

\bibitem[{Ngiam et~al(2021)Ngiam, Caine, Vasudevan, Zhang, Chiang et~al}]{2021scene-transformer}
Ngiam J, Caine B, Vasudevan V, et~al (2021) Scene transformer: A unified multi-task model for behavior prediction and planning. arXiv preprint arXiv:210608417 2(7)

\bibitem[{Park et~al(2024)Park, Jeong, Yoon, Jeong, and Yoon}]{park2024t4p}
Park D, Jeong J, Yoon SH, et~al (2024) T4p: Test-time training of trajectory prediction via masked autoencoder and actor-specific token memory. In: Proceedings of the IEEE/CVF Conference on Computer Vision and Pattern Recognition, pp 15065--15076

\bibitem[{Peri et~al(2022)Peri, Luiten, Li, Ošep, Leal-Taixé, and Ramanan}]{peri2022forecastinglidarfutureobject}
Peri N, Luiten J, Li M, et~al (2022) Forecasting from lidar via future object detection. \urlprefix\url{https://arxiv.org/abs/2203.16297}, \eprint{2203.16297}

\bibitem[{Phan-Minh et~al(2020)Phan-Minh, Grigore, Boulton, Beijbom, and Wolff}]{covernet}
Phan-Minh T, Grigore EC, Boulton FA, et~al (2020) Covernet: Multimodal behavior prediction using trajectory sets. In: Proceedings of the IEEE/CVF conference on computer vision and pattern recognition, pp 14074--14083

\bibitem[{Polychronopoulos et~al(2007)Polychronopoulos, Tsogas, Amditis, and Andreone}]{polychronopoulos2007sensor}
Polychronopoulos A, Tsogas M, Amditis AJ, et~al (2007) Sensor fusion for predicting vehicles' path for collision avoidance systems. IEEE Transactions on Intelligent Transportation Systems 8(3):549--562

\bibitem[{Qingkai et~al(2020)Qingkai, Manjiang, Guotao, and Xiaowei}]{qingkai2020lightweight}
Qingkai W, Manjiang H, Guotao X, et~al (2020) Lightweight hd map construction for autonomous vehicles in non-paved roads. Tech. rep., SAE Technical Paper

\bibitem[{Salzmann et~al(2020)Salzmann, Ivanovic, Chakravarty, and Pavone}]{salzmann2020trajectron++}
Salzmann T, Ivanovic B, Chakravarty P, et~al (2020) Trajectron++: Dynamically-feasible trajectory forecasting with heterogeneous data. In: Computer Vision--ECCV 2020: 16th European Conference, Glasgow, UK, August 23--28, 2020, Proceedings, Part XVIII 16, Springer, pp 683--700

\bibitem[{Sch{\"o}ller et~al(2020)Sch{\"o}ller, Aravantinos, Lay, and Knoll}]{scholler2020constant}
Sch{\"o}ller C, Aravantinos V, Lay F, et~al (2020) What the constant velocity model can teach us about pedestrian motion prediction. IEEE Robotics and Automation Letters 5(2):1696--1703

\bibitem[{Seff et~al(2023)Seff, Cera, Chen, Ng, Zhou, Nayakanti, Refaat, Al-Rfou, and Sapp}]{seff2023motionlm}
Seff A, Cera B, Chen D, et~al (2023) Motionlm: Multi-agent motion forecasting as language modeling. In: Proceedings of the IEEE/CVF International Conference on Computer Vision, pp 8579--8590

\bibitem[{Shi et~al(2022)Shi, Jiang, Dai, and Schiele}]{MTR}
Shi S, Jiang L, Dai D, et~al (2022) Motion transformer with global intention localization and local movement refinement. Advances in Neural Information Processing Systems 35:6531--6543

\bibitem[{Shi et~al(2024)Shi, Jiang, Dai, and Schiele}]{shi2024mtr++}
Shi S, Jiang L, Dai D, et~al (2024) Mtr++: Multi-agent motion prediction with symmetric scene modeling and guided intention querying. IEEE Transactions on Pattern Analysis and Machine Intelligence

\bibitem[{Sun et~al(2022)Sun, Huang, Gu, Williams, and Zhao}]{sun2022m2i}
Sun Q, Huang X, Gu J, et~al (2022) M2i: From factored marginal trajectory prediction to interactive prediction. In: Proceedings of the IEEE/CVF Conference on Computer Vision and Pattern Recognition, pp 6543--6552

\bibitem[{Sun et~al(2024)Sun, Zhang, Ma, Shi, Li, Luo, Wang, Xu, Cao, and Zhao}]{sun2024largetrajectorymodelsscalable}
Sun Q, Zhang S, Ma D, et~al (2024) Large trajectory models are scalable motion predictors and planners. \urlprefix\url{https://arxiv.org/abs/2310.19620}, \eprint{2310.19620}

\bibitem[{Teeti et~al(2022)Teeti, Khan, Shahbaz, Bradley, and Cuzzolin}]{teeti2022vision-based-survey}
Teeti I, Khan S, Shahbaz A, et~al (2022) Vision-based intention and trajectory prediction in autonomous vehicles: A survey. In: Proceedings of the Thirty-First International Joint Conference on Artificial Intelligence, IJCAI-22, Lud De Raedt, Ed, pp 5630--5637

\bibitem[{Varadarajan et~al(2022)Varadarajan, Hefny, Srivastava, Refaat, Nayakanti et~al}]{multipath++}
Varadarajan B, Hefny A, Srivastava A, et~al (2022) Multipath++: Efficient information fusion and trajectory aggregation for behavior prediction. In: 2022 International Conference on Robotics and Automation (ICRA), IEEE, pp 7814--7821

\bibitem[{Wagner et~al(2023)Wagner, Klemp, Lopez, and Tas}]{Road-Barlow}
Wagner R, Klemp M, Lopez CF, et~al (2023) Road barlow twins: Redundancy reduction for motion prediction. In: ICRA2023 Workshop on Pretraining for Robotics (PT4R)

\bibitem[{Wang et~al(2024)Wang, Messaoud, Liu, Gall, and Alahi}]{wang2024forecast}
Wang J, Messaoud K, Liu Y, et~al (2024) Forecast-peft: Parameter-efficient fine-tuning for pre-trained motion forecasting models. arXiv preprint arXiv:240719564

\bibitem[{Wang et~al(2023)Wang, Chen, Cheng, Mei, Song, and Liu}]{POP}
Wang S, Chen Y, Cheng J, et~al (2023) Improving autonomous driving safety with pop: A framework for accurate partially observed trajectory predictions. arXiv preprint arXiv:230915685

\bibitem[{Weng et~al(2022)Weng, Ivanovic, Kitani, and Pavone}]{weng2022affinipred}
Weng X, Ivanovic B, Kitani K, et~al (2022) Whose track is it anyway? improving robustness to tracking errors with affinity-based trajectory prediction. In: 2022 IEEE/CVF Conference on Computer Vision and Pattern Recognition (CVPR), pp 6563--6572, \doi{10.1109/CVPR52688.2022.00646}

\bibitem[{Wilson et~al(2023)Wilson, Qi, Agarwal, Lambert, Singh, Khandelwal, Pan, Kumar et~al}]{2023argoverse2}
Wilson B, Qi W, Agarwal T, et~al (2023) Argoverse 2: Next generation datasets for self-driving perception and forecasting. arXiv preprint arXiv:230100493

\bibitem[{Xie et~al(2017)Xie, Gao, Qian, Huang, Li, and Wang}]{xie2017vehicle}
Xie G, Gao H, Qian L, et~al (2017) Vehicle trajectory prediction by integrating physics-and maneuver-based approaches using interactive multiple models. IEEE Transactions on Industrial Electronics 65(7):5999--6008

\bibitem[{Xu et~al(2022)Xu, Li, Tang, Sun, Keutzer, Tomizuka, Fathi, and Zhan}]{Pretram}
Xu C, Li T, Tang C, et~al (2022) Pretram: Self-supervised pre-training via connecting trajectory and map. In: European Conference on Computer Vision, Springer, pp 34--50

\bibitem[{Xu et~al(2024)Xu, Chambon, Éloi Zablocki, Chen, Alahi, Cord, and Pérez}]{xu2024motionforecastingrealworldperception}
Xu Y, Chambon L, Éloi Zablocki, et~al (2024) Towards motion forecasting with real-world perception inputs: Are end-to-end approaches competitive? \urlprefix\url{https://arxiv.org/abs/2306.09281}, \eprint{2306.09281}

\bibitem[{Yuan et~al(2021)Yuan, Weng, Ou, and Kitani}]{2021agentformer}
Yuan Y, Weng X, Ou Y, et~al (2021) Agentformer: Agent-aware transformers for socio-temporal multi-agent forecasting. In: Proceedings of the IEEE/CVF International Conference on Computer Vision, pp 9813--9823

\bibitem[{Zeng et~al(2019)Zeng, Luo, Suo, Sadat, Yang, Casas, and Urtasun}]{zeng2019end}
Zeng W, Luo W, Suo S, et~al (2019) End-to-end interpretable neural motion planner. In: Proceedings of the IEEE/CVF Conference on Computer Vision and Pattern Recognition, pp 8660--8669

\bibitem[{Zeng et~al(2021)Zeng, Liang, Liao, and Urtasun}]{lanercnn}
Zeng W, Liang M, Liao R, et~al (2021) Lanercnn: Distributed representations for graph-centric motion forecasting. In: 2021 IEEE/RSJ International Conference on Intelligent Robots and Systems (IROS), IEEE, pp 532--539

\bibitem[{Zhan et~al(2019)Zhan, Sun, Wang, Shi, Clausse et~al}]{zhan2019interaction}
Zhan W, Sun L, Wang D, et~al (2019) Interaction dataset: An international, adversarial and cooperative motion dataset in interactive driving scenarios with semantic maps. \eprint{1910.03088}

\bibitem[{Zhang et~al(2022)Zhang, Bai, Xue, Fang, Zheng, and Ouyang}]{zhang2022trajectoryforecastingdetectionuncertaintyaware}
Zhang P, Bai L, Xue J, et~al (2022) Trajectory forecasting from detection with uncertainty-aware motion encoding. \urlprefix\url{https://arxiv.org/abs/2202.01478}, \eprint{2202.01478}

\bibitem[{Zhao et~al(2021)Zhao, Gao, Lan, Sun, Sapp, Varadarajan, Shen, Shen, Chai, Schmid et~al}]{tnt}
Zhao H, Gao J, Lan T, et~al (2021) Tnt: Target-driven trajectory prediction. In: Conference on Robot Learning, PMLR, pp 895--904

\bibitem[{Zhou et~al(2024)Zhou, Shao, Wang, Waslander, Li, and Liu}]{zhou2024smartrefine}
Zhou Y, Shao H, Wang L, et~al (2024) Smartrefine: A scenario-adaptive refinement framework for efficient motion prediction. In: Proceedings of the IEEE/CVF Conference on Computer Vision and Pattern Recognition, pp 15281--15290

\bibitem[{Zhou et~al(2022)Zhou, Ye, Wang, Wu, and Lu}]{2022hivt}
Zhou Z, Ye L, Wang J, et~al (2022) Hivt: Hierarchical vector transformer for multi-agent motion prediction. In: Proceedings of the IEEE/CVF Conference on Computer Vision and Pattern Recognition, pp 8823--8833

\bibitem[{Zhou et~al(2023{\natexlab{a}})Zhou, Wang, Li, and Huang}]{qcnet}
Zhou Z, Wang J, Li YH, et~al (2023{\natexlab{a}}) Query-centric trajectory prediction. In: Proceedings of the IEEE/CVF Conference on Computer Vision and Pattern Recognition, pp 17863--17873

\bibitem[{Zhou et~al(2023{\natexlab{b}})Zhou, Wen, Wang, Li, and Huang}]{qcnext}
Zhou Z, Wen Z, Wang J, et~al (2023{\natexlab{b}}) Qcnext: A next-generation framework for joint multi-agent trajectory prediction. arXiv preprint arXiv:230610508

\end{thebibliography}

\end{document}